\documentclass[runningheads]{llncs}

\usepackage{eccv}

\usepackage{eccvabbrv}

\usepackage{graphicx}
\usepackage{booktabs}

\usepackage[accsupp]{axessibility}  

\usepackage{hyperref}

\usepackage{orcidlink}

\usepackage{xcolor,colortbl}
\usepackage{multirow}
\usepackage{amsmath}
\usepackage{amssymb}
\usepackage{dsfont}
\usepackage{pifont}
\usepackage[misc]{ifsym}

\usepackage[ruled,vlined,linesnumbered]{algorithm2e}
\makeatletter
\newcommand{\algorithmfootnote}[2][\footnotesize]{%
  \let\old@algocf@finish\@algocf@finish
  \def\@algocf@finish{\old@algocf@finish
    \leavevmode\rlap{\begin{minipage}{\linewidth}
    #1#2
    \end{minipage}}%
  }%
}

\newcommand{\indic}[1]{\mathds{1}_{\{#1\}}}
\newcommand{\boldparagraph}[1]{\vspace{0.cm}\noindent{\bf #1}}
\newcommand{\cmark}{\ding{51}}%
\newcommand{\xmark}{\ding{55}}%
\begin{document}

\title{Lane Graph as Path: \\Continuity-preserving Path-wise  Modeling for Online Lane Graph Construction} 

\titlerunning{LaneGAP}

\author{Bencheng Liao\inst{1,2,\star} \and
Shaoyu Chen\inst{2,\star} \and
Bo Jiang\inst{2} \and
Tianheng Cheng\inst{2} \and
Qian Zhang\inst{3} \and
Wenyu Liu\inst{2} \and
Chang Huang\inst{3} \and
Xinggang Wang\inst{2,\textrm{\Letter}}
}

\authorrunning{Liao et al.}

\institute{Institute of Artificial Intelligence, Huazhong University of Science \& Technology \and
School of EIC, Huazhong University of Science \& Technology \and
Horizon Robotics}

\maketitle
\let\thefootnote\relax\footnotetext{$^\star$ Equal contribution; $^\textrm{\Letter}$ Corresponding author: \texttt{xgwang@hust.edu.cn}}
\begin{abstract}
Online lane graph construction is a  promising but challenging task in autonomous driving. Previous methods usually model the lane graph at the pixel or piece level, and recover the lane graph by pixel-wise or piece-wise connection, which breaks down the continuity of the lane and results in suboptimal performance. Human drivers focus on and drive along the continuous and complete paths instead of considering lane pieces. Autonomous vehicles also require path-specific guidance from lane graph for trajectory planning. We argue that the path, which indicates the traffic flow, is the primitive of the lane graph. Motivated by this, we propose to model the lane graph in a novel path-wise manner,  which well preserves the continuity of the lane and encodes traffic information for planning. We present a path-based online lane graph construction method, termed LaneGAP, which end-to-end learns the path and recovers the lane graph via a Path2Graph algorithm. We qualitatively and quantitatively demonstrate the superior accuracy and efficiency of LaneGAP over conventional pixel-based and piece-based methods on the challenging nuScenes and Argoverse2 datasets under controllable and fair conditions. Compared to the recent state-of-the-art piece-wise method TopoNet on the OpenLane-V2 dataset, LaneGAP still outperforms by 1.6 mIoU, further validating the effectiveness of path-wise modeling.
Abundant visualizations in the supplementary material show LaneGAP can cope with diverse traffic conditions.
Code is released at \url{https://github.com/hustvl/LaneGAP}.
   
\end{abstract}

\section{Introduction}
\label{sec:intro}

\begin{figure}[t!]
    \centering
    \includegraphics[width=\linewidth]{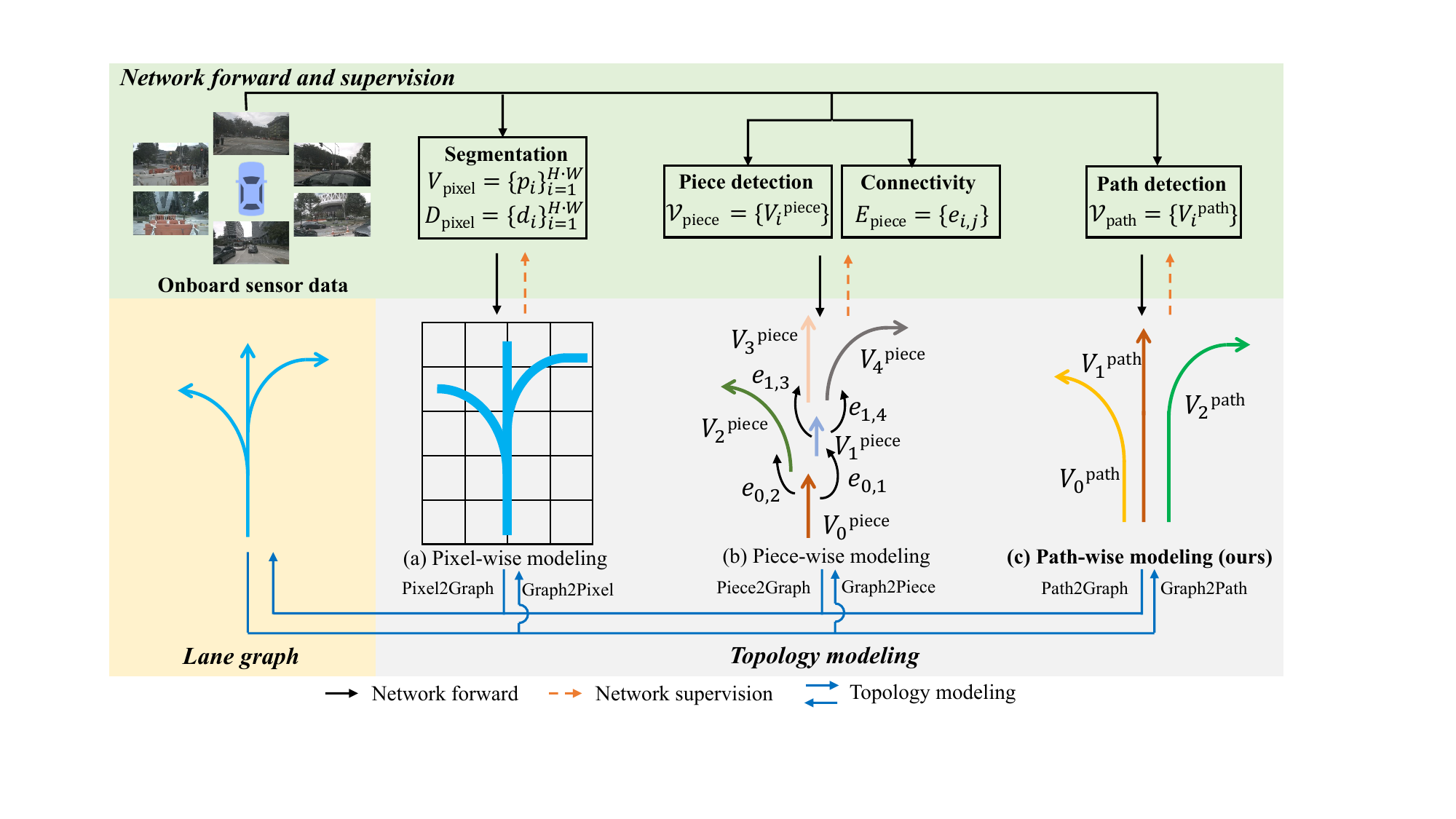}

    \caption{\textbf{Modeling comparison.} \textbf{(a)} Pixel-wise modeling~\cite{hdmapnet} utilizes a predefined Graph2Pixel algorithm to rasterize the lane graph into a segmentation map and a direction map on dense BEV pixels, and heuristic Pixel2Graph post-processing is needed to recover the lane graph from the predicted segmentation map $V_{\text{pixel}}$ and direction map $D_{\text{pixel}}$ (direction map is not drawn here for simplicity). \textbf{(b)} Piece-wise modeling~\cite{stsu} utilizes a predefined Graph2Piece algorithm to  split the lane graph into a set of pieces and  the connectivity matrix among pieces, and then it merges the predicted pieces $\mathcal{V}_{\text{piece}}$ to the graph with the Piece2Graph algorithm based on predicted connectivity $E_{\text{piece}} $. \textbf{(c)} The proposed path-wise modeling translates the lane graph into complete paths with a predefined Graph2Path algorithm to traverse the graph. 
    We perform path detection
    and adopt a Path2Graph algorithm  to recover the lane graph.}
    \label{fig:topo_model}

    \end{figure}

The lane graph  contains detailed lane-level traffic information,
and serves to provide path-specific guidance for trajectory planning, \ie, an automated vehicle can trace a path from the lane graph as reliable planning prior.

Lane graph is traditionally constructed with an offline map generation pipeline.  
However, autonomous driving demands a high degree of freshness in lane topology. Hence, online lane graph construction with vehicle-mounted sensors (\eg, cameras and LiDAR) is of great application value.





\begin{figure}[t!]
    \centering
    \includegraphics[width=0.9\linewidth]{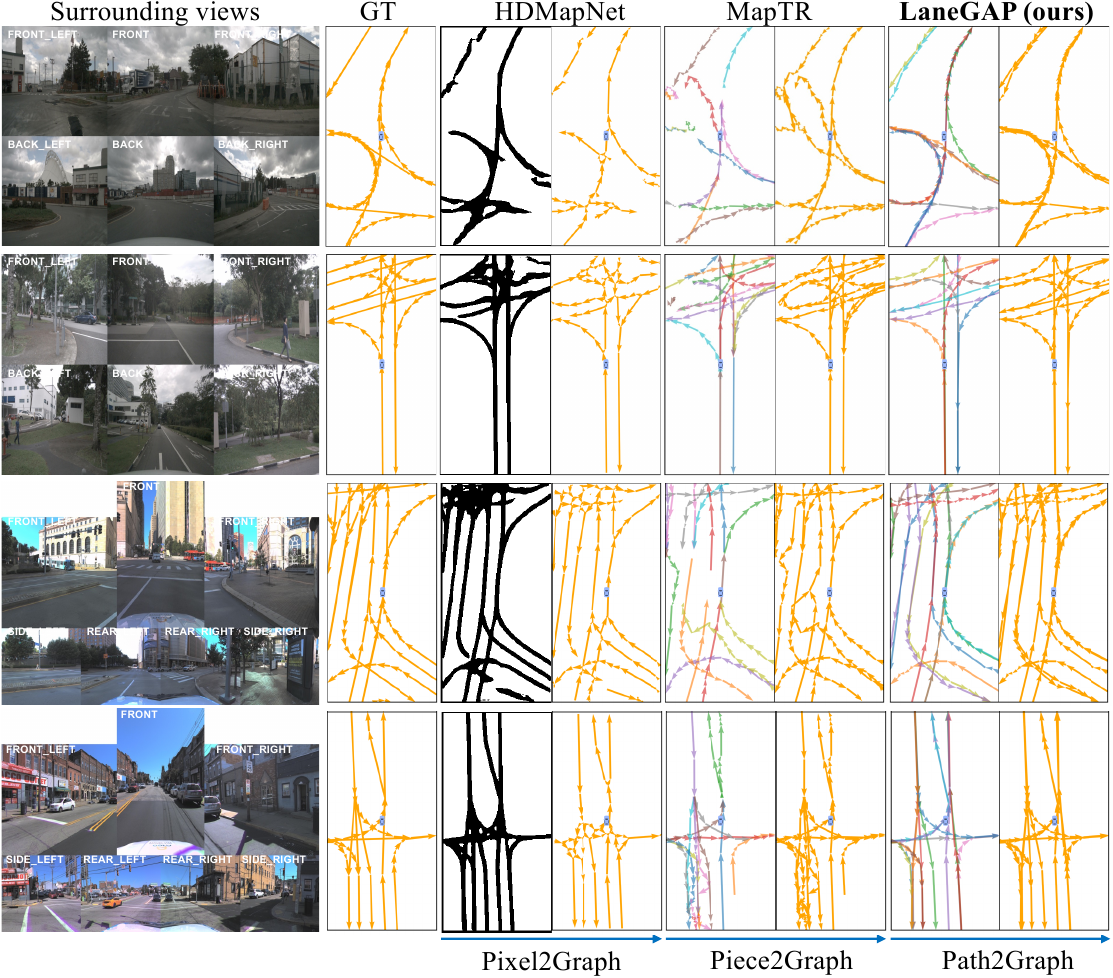}
    \caption{\textbf{Qualitative comparison of pixel-wise HDMapNet, piece-wise MapTR, and path-wise LaneGAP on the nuScenes and Argoverse2 val splits.} The top 2 rows are from the nuScenes val split, and the bottom 2 rows are from the Argoverse2 val split. Different colors indicate different instances. More qualitative comparisons are available in the supplementary.}
    \label{fig:qualitative_comparison}

    \end{figure}

An intuitive solution is to model the lane graph in a pixel-wise manner and adopt a segmentation-then-vectorization paradigm.
For example,  HDMapNet~\cite{hdmapnet} predicts a  segmentation map and a direction map on dense bird's-eye-view (BEV) features.
It then extracts the skeleton from the coarse segmentation map with a morphological thinning algorithm and extracts the graph topology by greedily tracing the single-pixel-width skeleton with the predicted direction map. 
Pixel-wise modeling incurs heuristic and time-consuming post-processing and often fails in complicated topology (see Fig.~\ref{fig:qualitative_comparison} row 3, Tab.~\ref{tab:postproc}).

Lane graph is also modeled in a piece-wise manner~\cite{stsu}, which splits the lane graph into lane pieces at junction points (\ie, merging points and fork points) and predicts an inter-piece connectivity matrix.  Based on the connectivity, pieces are linked and merged into the lane graph. 
However, piece-wise modeling breaks down the continuity of the lane. At complicated road intersections,
the fragmented short pieces are hard for a neural network to learn to detect (\eg, $V_1^{\text{piece}}$ in Fig.~\ref{fig:topo_model} (b)), let alone predict the connections.

We argue that the path is the primitive of the lane graph.
Human drivers focus on and drive along the continuous 
 and complete paths instead of considering lane pieces. 
Autonomous vehicles also require path-specific guidance from 
 lane graph for trajectory planning~\cite{Dauner2023CORL}.
Continuous paths play an important role in indicating the driving flow, while pixel-wise and piece-wise modeling often fail to merge pixels and pieces into continuous paths (see Fig.~\ref{fig:qualitative_comparison}).

Motivated by this, we propose to model the lane graph in an alternative path-wise manner. 
We decouple the lane graph into a set of continuous paths with a proposed Graph2Path algorithm,  perform path detection through set prediction~\cite{detr, maptr}, 
and extract a fine-grained lane graph with a Path2Graph algorithm.  
Based on this path-wise modeling, we propose an online lane graph construction framework, termed LaneGAP, which  feeds onboard sensor data into an end-to-end network for path detection, and further transforms detected paths into a lane graph.

To fairly evaluate across different modelings and emphasize the unique properties of the lane graph (\ie, connection, direction, and split/merge around the junction points), we carefully design a graph-centric TOPO metric to evaluate both holistic topology and local junction topology, which is compatible with different modelings.

We compare LaneGAP with the pixel-wise modeling method HDMapNet~\cite{hdmapnet} and piece-wise modeling methods STSU~\cite{stsu}, MapTR~\cite{maptr} under strictly fair conditions (encoder, model size, training schedule, \etc) on the challenging nuScenes~\cite{nuscenes} and Argoverse2~\cite{av2} datasets, which covers diverse graph topology and traffic conditions. With only camera input, LaneGAP achieves the best graph construction quality both quantitatively (see Tab.~\ref{tab:modeling}) and qualitatively (see Fig.~\ref{fig:qualitative_comparison}), while running at the fastest inference speed (2x faster than HDMapNet). 
We further validate the superiority of path-wise modeling on TopoNet~\cite{toponet} (1.6 higher mIoU), a state-of-the-art piece-wise modeling method on the OpenLane-V2 dataset~\cite{openlanev2}.
The extensive experiments show that modeling the lane graph at the path level is  reasonable and promising. 
We hope LaneGAP can serve as a fundamental module of the self-driving system and boost the development of downstream motion planning.

Our contributions can be summarized as follows:
\begin{itemize}
    \item  
    We propose to model the lane graph in a novel path-wise manner,  which well preserves the continuity of the lane and encodes traffic information for planning.
    \item  Based on our path-wise modeling, we present an online lane graph construction method, termed LaneGAP. LaneGAP end-to-end learns the path and constructs the lane graph via the designed Path2Graph algorithm.
    \item We design a novel graph-centric metric to evaluate the unique properties of lane graph construction (\ie, connection, direction, and junction), which is compatible with different modelings. 
    \item 
    We  qualitatively and quantitatively demonstrate the superior accuracy and efficiency of LaneGAP over pixel-based and piece-based methods. 
    LaneGAP can cope with diverse traffic conditions, especially for road intersections with complicated lane topology.
\end{itemize}

\section{Related Work}
\label{sec:relatedwork}

\boldparagraph{Lane detection.}
Lane detection only considers predicting and evaluating the lane divider lines without spatial relation (merging and fork). 
Since most lane detection datasets only provide front-view images, previous lane detection methods~\cite{tabelini2021keep,wang2022keypoint,garnett20193d,lstr,guo2020gen,liu2022learning} were stuck in predicting lines with a small curvature in a limited horizontal FOV. BezierLaneNet~\cite{feng2022rethinking} uses a fully convolutional network to predict Bezier lanes defined with 4 Bezier control points. PersFormer~\cite{chen2022persformer} 
proposes a Transformer-based architecture for spatial transformation and unifies 2D and 3D lane detection. 

\boldparagraph{Online HD map construction.}
Online HD map construction can be seen as an advanced setting of lane detection, consisting of  lines and polygons with various semantics in the local 360$^\circ$ FOV  perception range of ego-vehicle.
With advanced 2D-to-BEV modules~\cite{Ma2022VisionCentricBP}, previous online HD map construction methods cast it into semantic segmentation task on the transformed BEV features~\cite{polarbev,cvt,bevformer,liu2022bevfusion,liu2022petrv2,lu2022ego3rt}. Building vectorized semantic HD map online achieves increasing interests nowadays~\cite{hdmapnet,maptr,vectormapnet,instagram,bemapnet,pivotnet}, HDMapNet~\cite{hdmapnet} follows a segmentation-then-vectorization paradigm.
To achieve end-to-end learning~\cite{detr,deformdetr,yolos}, VectorMapNet~\cite{vectormapnet} 
adopts a  two-stage pipeline for vectorized HD map learning.
MapTR~\cite{maptr} proposes a unified permutation-equivalent modeling to exploit the undirected nature of semantic HD map and designs a parallel end-to-end framework. BeMapNet~\cite{bemapnet} and PivotNet~\cite{pivotnet} propose  Bezier-based representation and pivot-based representation for modeling map geometry. While the above works focus on map elements without physical directions and lane graph topology, we aim to fill the gap in this work.

\boldparagraph{Road graph construction.}
There is a long history of extracting the road graph from remote sensor data (\eg, aerial imagery and satellite imagery). 
Many works~\cite{Mattyus_2017_ICCV, zhou2018d, batra2019improved,He2020Sat2GraphRG,buslaev2018fully}
frame the road graph as a pixel-wise segmentation problem  
and utilizes morphological post-processing methods to extract the road graph. RoadTracer~\cite{bastani2018roadtracer} uses an iterative search process to extract graph topology step by step. Some works~\cite{chu2019neural,Tan_2020_CVPR,xu2021icurb,li2019topological,mi2021hdmapgen} follow this sequential generation paradigm. Different from the above works, we focus on the online, ego-centric setting with vehicle-mounted sensors to produce more fine-grained lane-level graph.

\boldparagraph{Lane graph construction.}
The lane graph is traditionally constructed with an offline pipeline~\cite{centerlinedet,laneextract, buchner2023learning}. \cite{laneextract} proposes a multi-step training pipeline to construct the lane graph of aerial images based on pixel-wise modeling. \cite{buchner2023learning} proposes a bottom-up approach to aggregate multiple local aerial lane graphs into a globally consistent graph. CenterlineDet~\cite{centerlinedet} proposes a DETR-like decision-making transformer network to iteratively update the global lane graph with vehicle-mounted sensors.
Recently, STSU~\cite{stsu} shifts the offline lane graph construction to the online, ego-centric setting with vehicle-mounted sensors. It models the lane graph as a set of disjoint pieces split by junction points and a set of connections among those pieces. 
Based on STSU, recent works~\cite{can2022topology,toponet,wu2024topomlp} advance this piece-wise setting and push forward the performance.  Different from the above graph modelings, we regard the path as the primitive of the lane graph, and model the lane graph in a novel path-wise manner.




\section{Method}

\label{sec:g2p}

\begin{figure}[t!]
    \centering
    \includegraphics[width=\linewidth]{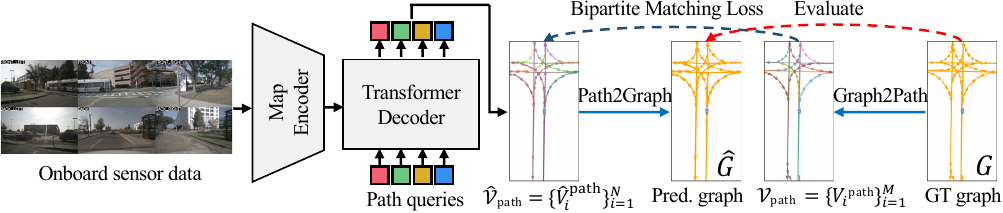}

    \caption{\textbf{Overview of LaneGAP.} }
    \label{fig:framework}

    \end{figure}

In this section, we first describe how to translate the directed lane graph into a set of directed paths in Sec.~\ref{sec:g2p}. Then we introduce the online path detection framework in Sec.~\ref{sec:lanegap}.  And we describe how to translate  paths back to the lane graph in Sec.~\ref{sec:p2g}. An overview of our method is exhibited in Fig.~\ref{fig:framework}.

\subsection{Graph2Path}
We propose a simple Graph2Path algorithm to translate the directed lane graph into a set of paths according to the direction and connection information encoded in the lane graph. The pseudo-code of Graph2Path is shown in Alg.~\ref{algo:g2p}.

Given the ground truth Lane graph $G$, which is typically a directed graph in the local map around the ego-vehicle, we first extract the root vertices $\Omega_{\text{root}}$ and leaf vertices $\Omega_{\text{leaf}}$. Then we pair the root vertices and leaf vertices and then utilize depth-first-search (DFS) algorithm to find the valid path $V^{\text{path}}$. Finally, we can translate the ground truth directed lane graph $G$ to a set of directed paths $\mathcal{V}_{\rm path} = \{V_i^{\rm path}\}_{i=1}^{M}$, where $M$ is the number of ground truth paths.

\subsection{Path Representation and Learning}
\label{sec:lanegap}
Inspired by  advanced set detection methods~\cite{detr,deformdetr}, we propose an end-to-end network, LaneGAP, to predict all paths simultaneously in a single stage, as illustrated in Fig.~\ref{fig:framework}. Our network consists of an encoder that encodes the features  from the onboard sensor data, and a query-based Transformer decoder that performs set detection by decoding a set of paths $\hat{\mathcal{V}}_{\rm path} = \{ \hat{V}_i^{\rm path}\}_{i=1}^N$ from the encoded features. To parameterize the path, we utilize two widely-used types of representations, Polyline~\cite{vectormapnet,maptr} and Bezier~\cite{stsu,bemapnet}, where Polyline offers high flexibility in describing the path, while Bezier provides a smoother representation.

\boldparagraph{Polyline representation.}
Polyline representation models the arbitrary directed path as an ordered set of $N_{\text{p}}$ points  $V_{\text{poly}}^{\text{path}} = \{p_j \in \mathbb{R}^2 | j = 0,1,2,..., N_{\text{p}}-1\}$. We directly regress polyline points and utilize deformable attention~\cite{deformdetr} to exploit the local information along each Polyline path, where the keys and values are the local features along the Polyline path.

\boldparagraph{Bezier representation.}
Bezier representation models the directed path as an ordered set of $N_{\text{b}}$ control points $V_{\text{Bezier}}^{\text{path}} = \{b_j \in \mathbb{R}^2 | j = 0,1,2,..., N_{\text{b}}-1\}$.
Bezier is a parametric curve, where the point $B$ on the line  can be  calculated by the weighted sum of control points $V_{\text{Bezier}}^{\text{path}}$:
\begin{equation}
\begin{gathered}
B = \sum_{j=0}^{N_{\text{b}}-1} C_{N_{\text{b}}-1}^{j}t^{j}(1-t)^{N_{\text{b}}-1-j} b_j, 0 \leq t \leq 1.
\end{gathered}   
\label{eq:1}
\end{equation}
Given the Bezier control points set $V_{\text{Bezier}}^{\text{path}}$ and the sampled interval set $T=\{ t_k \in \mathbb{R}|0 \leq t_k \leq 1, k=0,1,2,...,K-1 \}$, we can calculate the curve $\mathcal{B} = \{B_{k}\in \mathbb{R}^2|0,1,2,...,K-1 \}$ with matrix multiplication:
\begin{equation}
\begin{gathered}
\mathcal{B} = \Gamma \times V_{\text{Bezier}}^{\text{path}},
\end{gathered}
\label{eq:2}
\end{equation}
where weight matrix $\Gamma$ is a $K\times N_{\text{b}}$ matrix and $\Gamma(k,j) = C_{N_{\text{b}}-1}^j t_{k}^{j} (1-t_k)^{N_{\text{b}}-1-j}$. For Beizer representation, we use the same network as Polyline representation to regress Bezier control points directly. To enable exploiting the local information along the Bezier path with offline control points, we sample the Bezier path to get online points $\mathcal{B}$ based on Eq.~\ref{eq:2} and perform deformable attention, where the keys and values are the local features around the sampled points $\mathcal{B}$ along the Bezier path. We denote this design as Bezier deformable attention, which enables the Transformer decoder to aggregate features along the Bezier path.

\boldparagraph{Learning.}
We use the encoder to transform the input onboard sensor data into unified BEV features $F_{\rm BEV}$~\cite{bevformer,gkt,second}. Then we utilize the Transformer decoder to predict a set of paths based on a set of random initialized learnable path queries~\cite{maptr}.
With the above path representation, we can cast the predicted path into an ordered set of points with a fixed number $N_v$ of points on the path, where $\hat{V}_i^{\text{path}} = \{\hat{v}_j \in \mathbb{R}^2|j=0,1,2,...,N_v-1\}$. We modify the bipartite matching loss used in~\cite{detr} to fit in the path detection setting:
\begin{equation}
\begin{aligned}
\mathcal{L}_{\rm bipartite}(\hat{\mathcal{V}}_{\text{path}},\mathcal{V}_{\text{path}}) =& \sum_{i=1}^{N}[\mathcal{L}_{\rm{Focal}}(\hat{p}_{{\hat{\sigma}}(i)}, c_i) + &\indic{c_i \neq \varnothing}\mathcal{L}_{\rm path}(\hat{V}_{\hat{\sigma}(i)}^{\rm path}, V_i^{\rm path})],\\
\mathcal{L}_{\rm path}(\hat{V}_{\hat{\sigma}(i)}^{\rm path}, V_i^{\rm path}) =& \sum_{j=0}^{N_v-1}L_1(\hat{v}_j,v_j),
\end{aligned}    
\label{eq:3}
\end{equation}
where $\hat{\sigma}$ is the optimal assignment between a set of predicted paths and a set of ground truth paths computed by the Hungarian algorithm, $c_i$ is the target class label, and $\mathcal{L}_{\rm{Focal}}(\hat{p}_{{\hat{\sigma}}(i)}, c_i)$ is the classification loss defined in~\cite{focal}. $L_1$ loss is utilized between the matched predicted path $\hat{V}_{\hat{\sigma}(i)}^{\rm path}$ and sampled ground truth path $V_i^{\rm path}$.
To enhance the BEV features, we introduce an auxiliary BEV segmentation branch $\phi_{\rm seg}$  to predict the BEV segmentation mask, and the auxiliary BEV segmentation loss is defined as:
\begin{equation}
\begin{gathered}
\mathcal{L_{\rm auxseg}} = \mathcal{L_{\rm CE}}(\phi_{\rm seg}(F_{\rm BEV}), M_{\rm BEV}),
\end{gathered}
\label{eq:4}
\end{equation}
where $M_{\rm BEV}$ is the ground truth. The total loss is defined as:
\begin{equation}
\begin{gathered}
\mathcal{L_{\rm total}} = \mathcal{L_{\rm cls}} + \mathcal{L_{\rm path}} + \mathcal{L_{\rm auxseg}},
\end{gathered}
\label{eq:5}
\end{equation}
where $\mathcal{L_{\rm cls}}$ is the path-wise classification loss using Focal loss~\cite{focal}.

\begin{minipage}[t]{0.43\textwidth}
    \begin{algorithm}[H]
    \SetAlgoLined
    \DontPrintSemicolon
    \SetNoFillComment
    \footnotesize
    \KwIn{directed lane graph $G$}
    \KwOut{directed paths $\mathcal{V}$}
    
    Initialize: $\mathcal{V} \leftarrow \emptyset$, $\text{root vertices set } \Omega_{\text{root}} \leftarrow \emptyset$, $\text{leaf vertices set } \Omega_{\text{leaf}} \leftarrow \emptyset$\;
    \For{v in $G$.vertices}{
        \If{ $\text{v}.\text{in\_degree} == 0$}{
            $\Omega_{\text{root}} \leftarrow \Omega_{\text{root}} \cup \{v\}$\;
        }
        \If{ $\text{v}.\text{out\_degree} == 0$}{
            $\Omega_{\text{leaf}} \leftarrow \Omega_{\text{leaf}} \cup \{v\}$\;
        }
    }
    \For{root v in $\Omega_{\text{root}}$ }{
        \For{leaf v in $\Omega_{\text{leaf}}$}{
            $V^{\text{path}} = \textbf{findpath}(root\ v, leaf\ v, G)$ \;
            \If{ $V^{\text{path}} \text{ is not None}$}{
                $\mathcal{V} \leftarrow \mathcal{V} \cup \{V^{\text{path}}\}$\;
            }
        }
    }
    
    Return: $\mathcal{V}$
    \caption{Pseudo-code of Graph2Path.}
    \label{algo:g2p}
    \end{algorithm}
    \end{minipage}
    \hfill
    \begin{minipage}[t]{0.49\textwidth}
    \begin{algorithm}[H]
    \SetAlgoLined
    \DontPrintSemicolon
    \SetNoFillComment
    \footnotesize
    \KwIn{a set of paths $\mathcal{V}=\{ V_i^{\text{path}} \} $}
    \KwOut{directed  lane graph $G$}
    
    Initialization: $G=\text{DirGraph}()$\;
    \For{$V^{\text{path}}$ in $\mathcal{V}$}{
        $V=\{v_j\} \leftarrow \textbf{discret\_to\_vert\_seq}(V^{\text{path}})$\;
        \For{ $v_j$ in $V$}{
            \If{ $v_j\ \text{is\ the\ last\ vertice\ of\ }V$}{
                $G.\textbf{add\_vertice}(v_j)$ \;
                \textbf{break}\;
            }
            $G.\textbf{add\_vertice}(v_j)$ \;
            $G.\textbf{add\_edge}(v_j,v_{j+1})$ \;
        }
    
    }
    $G=\textbf{merge\_vertices\_by\_overlap}(G)$

    Return: $G$
    \caption{Pseudo-code of Path2Graph.}
    \algorithmfootnote{The for loops here can be implemented in a parallel paradigm. $\textbf{merge\_vertices\_by\_overlap}$ merges the overlapped vertices of different paths to reduce the redundancy of $G$ without affecting the accuracy of graph topology.}
    \label{algo:p2g}
    \end{algorithm}
    \end{minipage}

\subsection{Path2Graph}
\label{sec:p2g}

The predicted continuous paths encode sufficient traffic information and can be directly applied to downstream motion planning. 
To further  recover the graph structure of lane topology and extract merging and fork information,
we convert the predicted paths $\hat{\mathcal{V}} = \{\hat{V}_i^{\text{path}}\}$ into a directed lane graph $\hat{G}$  with a  designed Path2Graph algorithm in Alg.~\ref{algo:p2g}. 

We discretize the path into point sequences $V=\{v_j\}$. The discretized points are regarded as vertices and the adjacent relation
between successive points is regarded as edges of vertices. We add these vertices and edges to the directed graph (line 4 to 11 in Alg.~\ref{algo:p2g}). 
The vertices registered in the directed graph $G$ on one path may have spatial overlapping with vertices on other paths. We merge the overlapped vertices into one vertex if their distance is below a threshold (line 13 in Alg.~\ref{algo:p2g}). The merged vertex inherits the predecessors and successors of the pair of overlapped vertices.

\section{Metrics}
Previous works mainly design a network based on one sort of graph modelings (pixel-wise or piece-wise) and choose a metric (pixel-level IoU or instance-level Chamfer distance mAP) friendly to the modeling. To fairly evaluate and compare across different modelings, we adapt the TOPO metric~\cite{laneextract} to measure the correctness of the overall directed graph construction. The uniqueness of lane graph construction lies in the junction points, without which the lane graph degenerates into a set of disjoint lanes. To emphasize the quality of the subgraph around the junction points, we propose a new metric, Junction TOPO, which specifically evaluates the accuracy of the local directed graph formed by traversing around the junction points on the directed lane graph.

\boldparagraph{TOPO metric.}
Given the predicted directed lane graph $\hat{G}$ and ground truth directed lane graph $G$, we interpolate them so that the distances between any two connected vertices are $0.15m$, and get predicted directed lane graph $\hat{G} = (\hat{V},\hat{E})$ and ground truth directed lane graph $G=(V,E)$ where $\hat{V},V$ are the sets of interpolated vertices and $\hat{E}, E $ are the sets of edges among vertices encoding direction and connection. For $\hat{V}$ and $V$,  a pair of vertices is considered a candidate match if the distance between the two vertices is less than $0.45m$. And  we utilize maximal one-to-one matching among those candidate pairs to find final matched vertices $P_{\text{pair}}=\{(\hat{v}, v)_{i}\}_{i=1}^{N_{\text{pair}}}$, then we traverse the directed graph around the paired vertices $\hat{v}$ and $v$ for less than $7.5m$ to get subgraphs $\hat{S}_{\hat{v}}$ and $S_v$  on $\hat{G}$ and G. We compute the precision $\text{Pre}(\hat{S}_{\hat{v}}, S_v) = \frac{N_{\text{subpair}}}{|\hat{S}_{\hat{v}}|}$and recall $\text{Rec}(\hat{S}_{\hat{v}},S_v) =  \frac{N_{\text{subpair}}}{|S_{v}|}$ between the vertices of predicted subgraph $\hat{S}_{\hat{v}}$ and the vertices of ground truth subgraph $S_{v}$, where the matching part follows the previous procedure with $0.45m$ threshold. Finally, we report the TOPO precision and recall defined as:
\begin{equation}
\begin{gathered}
\text{Precision}_{\text{TOPO}} = \frac{\sum_{i=1}^{N_{\text{pair}}} \text{Pre}(\hat{S}_{\hat{v}}, S_v)}{|\hat{V}|}, \\
\text{Recall}_{\text{TOPO}} = \frac{\sum_{i=1}^{N_{\text{pair}}} \text{Rec}(\hat{S}_{\hat{v}}, S_v)}{|V|}.
\end{gathered}    
\end{equation}

\boldparagraph{Junction TOPO metric.}
The TOPO metric focuses on the topology correctness of the overall directed lane graph, it does not highlight the correctness of the subgraph formed by traversing from the junction points, which plays a key role in determining the driving choices across different lanes. To bridge this gap, we propose the Junction TOPO metric, which only reports the precision and recall of the junction subgraph. Given the $N_{\text{junction}}$ junction points of the ground truth lane graph, we get pairs of subgraphs $( \hat{S}_{\text{junction}}, S_{\text{junction}})$ by traversing the directed graphs $\hat{G}$ and $G$ less than $7.5m$ from junction point. For each subgraph pair, we calculate the precision $\text{Pre}(\hat{S}_{\text{junction}}, S_{\text{junction}})$ and recall $ \text{Rec}(\hat{S}_{\text{junction}}, S_{\text{junction}})$.

\boldparagraph{Undirected versions.}
The above metrics calculate the precision and recall by traversing the directed graph $\hat{G}$ and $G$, ignoring the predecessor vertices. To evaluate the complete connections, we turn the directed graphs into undirected graphs and repeat the calculation defined above for two metrics.

\section{Experiments}

\subsection{Dataset}

We conduct experiments on two popular and large-scale datasets, \ie, nuScenes \cite{nuscenes} and Argoverse2~\cite{av2}. nuScenes~\cite{nuscenes} dataset consists of 1000 scene sequences. Each sequence is sampled in 2Hz frame rate and provides LiDAR point cloud and  RGB images from 6 surrounding cameras, which covers 360$^\circ$ horizontal FOV of the ego-vehicle.
The dataset provides the 2D lane graph without height information in the form of lane centerline and covers diverse online driving conditions. For the online setting of lane graph construction, we set the perception ranges as $[-15.0m, 15.0m]$ for the $X$-axis and $[-30.0m, 30.0m]$ for the $Y$-axis, and preprocess the dataset following~\cite{hdmapnet,vectormapnet,maptr}. We train on the nuScenes train set and evaluate on the val set.  The experiments on nuScenes dataset are conducted using 6 surrounding-view images by default.
Argoverse2~\cite{av2} dataset contains 1000 scene logs and 7 surrounding cameras with 360$^\circ$ horizontal FOV. Each log is sampled in 10Hz frame rate and provides 3D lane graph. Due to the limitations of pixel-wise modeling in performing 3D segmentation (piece-wise and path-wise modelings can simply predict an extra z-coordinate to support 3D lane graph construction), we omit the height information for fair comparison across different modelings. Other experimental settings remain the same as those for nuScenes.



\begin{table}[t!]
    \setlength{\tabcolsep}{2pt}
    \renewcommand\arraystretch{1.1}
    \centering
    \resizebox{0.98\textwidth}{!}{
    \begin{tabular}{l l  c  c  c  c  c  c  c  c  c  c  c  c  c c  c   c}
        \toprule
        \multirow{3}{*}{Dataset} & \multirow{3}{*}{Method} & \multirow{3}{*}{Modeling} & \multicolumn{6}{c}{Directed Graph} & \multicolumn{6}{c}{Undirected Graph} & \multirow{3}{*}{Param.} & \multirow{3}{*}{FPS$_{\text{net}}$} &\cellcolor{gray!20}\\ 
        
        & &  & \multicolumn{3}{c}{Junction TOPO} & \multicolumn{3}{c}{TOPO} 
         & \multicolumn{3}{c}{Junction TOPO} & \multicolumn{3}{c}{TOPO}  & & & \cellcolor{gray!20}\\ 
        & & & Prec. & Rec. & \cellcolor{gray!20}$F_1$ & Prec. & Rec. &  \cellcolor{gray!20}$F_1$
         & Prec. & Rec. & \cellcolor{gray!20}$F_1$  & Prec. & Rec. &  \cellcolor{gray!20}$F_1$ & & & \multirow{-3}{*}{\cellcolor{gray!20}FPS}\\
        \midrule
        \multirow{4}{*}{nusc}& HDMapNet & Pixel-wise  & 0.594 & 0.442 & \cellcolor{gray!20}0.507 & 0.608 & 0.443 & \cellcolor{gray!20}0.513 & 0.526 & 0.445 & \cellcolor{gray!20}0.482 & 0.563 & 0.447 &  \cellcolor{gray!20}0.498
        & 35.4M & 17.8 & \cellcolor{gray!20}6.8\\
        & STSU & Piece-wise & 0.449 & 0.393 & \cellcolor{gray!20}0.419 & 0.416 & 0.376 & \cellcolor{gray!20}0.395 & 0.413 & 0.404 & \cellcolor{gray!20}0.408 & 0.405 & 0.375 & \cellcolor{gray!20}0.389 & 35.3M & 15.8 &  \cellcolor{gray!20}15.1\\
        & MapTR & Piece-wise & 0.540 & 0.459 & \cellcolor{gray!20}0.496 & 0.504 & 0.432 &  \cellcolor{gray!20}0.465 & 0.493 & 0.465 &  \cellcolor{gray!20}0.478 & 0.487 & 0.429 &  \cellcolor{gray!20}0.456
        & 36.0M & 15.6 & \cellcolor{gray!20}15.0\\
        & LaneGAP & Path-wise & 0.591 & 0.539 & \cellcolor{gray!20}\textbf{0.564} & 0.547 & 0.511 & \cellcolor{gray!20}\textbf{0.529} & 0.543 & 0.518 & \cellcolor{gray!20}\textbf{0.530} & 0.534 & 0.496 & \cellcolor{gray!20}\textbf{0.514} & 35.9M & 16.5 & \cellcolor{gray!20}\textbf{15.6} \\
        \midrule
        \multirow{4}{*}{av2} & HDMapNet &Pixel-wise & 0.656 & 0.449 & \cellcolor{gray!20}0.533 & 0.635 & 0.463 & \cellcolor{gray!20}0.535 & 0.596& 0.473&\cellcolor{gray!20}0.528&0.607&0.470&\cellcolor{gray!20}0.530&35.4M&15.2& \cellcolor{gray!20}4.6\\
        & STSU & Piece-wise & 0.562&0.414&\cellcolor{gray!20}0.477&0.462&0.436&\cellcolor{gray!20}0.449&0.512&0.416&\cellcolor{gray!20}0.459&0.449&0.431&\cellcolor{gray!20}0.440&35.3M&13.0& \cellcolor{gray!20}12.4\\
        & MapTR & Piece-wise & 0.589&0.469&\cellcolor{gray!20}0.522&0.520&0.481&\cellcolor{gray!20}0.500&0.543&0.468&\cellcolor{gray!20}0.503&0.507&0.477&\cellcolor{gray!20}0.491&36.0M&13.1&  \cellcolor{gray!20}12.5\\
        & LaneGAP & Path-wise &0.663&0.575&\cellcolor{gray!20}\textbf{0.616}&0.578&0.552&\cellcolor{gray!20}\textbf{0.565} &0.627&0.541&\cellcolor{gray!20}\textbf{0.581}&0.569&0.546&\cellcolor{gray!20}\textbf{0.557}&35.9M&13.8& \cellcolor{gray!20}\textbf{13.0}\\
        \bottomrule
    \end{tabular}}

\caption{\textbf{Quantitative comparison of lane graph modeling under fair conditions} (the same encoder, comparable model size, the same training schedule, and vision-only modality). FPS$_{\text{net}}$ and FPS of all the experiments are measured on the same machine with one NVIDIA Geforce RTX 3090 GPU and one 24-core AMD EPYC 7402 2.8 GHz CPU, where FPS$_{\text{net}}$ is benchmarked with only network forward.}

\label{tab:modeling}

\end{table}
\begin{table}[t!]
    \centering
    \renewcommand\tabcolsep{3pt}
    \resizebox{0.98\linewidth}{!}{
    \begin{tabular}{l|ccccc|c}
        \toprule
        Method    &HDMapNet$^{*}_{\text{pixel}}$ & STSU$^{*}_{\text{piece}}$ & VectorMapNet$^{*}_{\text{piece}}$ & MapTR$^{*}_{\text{piece}}$ & TopoNet$^{*}_{\text{piece}}$ & LaneGAP$^{\dagger}$ \\
        \midrule
        mIoU  &  18.3 & 31.1 & 25.0 & 35.7&39.0&\textbf{40.6}\\
        \bottomrule
    \end{tabular}}
    \caption{\textbf{Quantitative comparison on Openlane-V2 dataset}. $^{*}$ means the results are directly taken from TopoNet paper~\cite{toponet}. LaneGAP$^{\dagger}$ is our proposed variant based on TopoNet$_{\text{piece}}$ to validate the effectiveness of path-wise modeling, which only adapts the lane piece detection part of TopoNet for learning path-wise representation.}
    \label{tab:toponet}

\end{table}

\subsection{Baselines}

\boldparagraph{Implementation details.}
All the baselines use the same encoder consisting of ResNet50~\cite{resnet} and GKT~\cite{gkt} to transform the  360$^\circ$ horizontal FOV multi-camera images to BEV features at the resolution of $200 \times 100$ (\ie, the BEV grid size is $0.3m$).   
All the models are controlled in comparable parameter sizes and trained for long enough epochs to ensure convergence (110 epochs on nuScenes and 24 epochs on Argoverse2). We train all the experiments on 8 NVIDIA GeForce RTX 3090 GPUs with a total batch size of 32 (6 view images on nuScenes and 7 view images on Argoverse2). 
Further details about training settings are provided in the supplementary material. 
The only differences lie in the specialized decoders designed for respective modelings, which are detailed as follows.


\boldparagraph{Pixel-wise modeling.}
We select HDMapNet~\cite{hdmapnet} as the pixel-wise modeling baseline. It adopts a U-Net structure with a ResNet18  to output a binarized segmentation map and a direction map with two branches. Meanwhile, a cross-entropy loss is applied to the segmentation map,  and  $L_2$ loss is applied to the (-1,1) normalized direction map~\cite{laneextract,zhou2018d}. To exclude the quantization error brought by classification on dense pixels, the output resolution of the U-Net-shaped~\cite{ronneberger2015u} segmentation model is set to $400 \times 200$, which means the grid size is $0.15m$ (the perception range is $60m\times30m$), the same as the graph interpolation size. The Pixel2Graph follows \cite{hdmapnet, laneextract}.



\boldparagraph{Piece-wise modeling.}
STSU~\cite{stsu} is selected as the piece-wise modeling baseline, which consists of a detection branch to detect Bezier lane pieces with 4 control points and an association branch to predict the connectivity between lane pieces. It stacks 6 Transformer decoder layers to iteratively refine the predictions. $L_1$ loss and Focal loss are applied to the matched (0,1) normalized pieces and cross-entropy loss is applied to the connectivity. More recently, MapTR~\cite{maptr} proposes a much stronger Transformer decoder for the detection branch. We further implement an additional piece-wise baseline by simply replacing the original Beizer piece detection branch of STSU with  MapTR Polyline piece detection branch. We set the number of Polyline points to 20. The Piece2Graph  follows \cite{stsu}.


\boldparagraph{Path-wise modeling.} 
The path-wise LaneGAP adopts the same decoder as the piece-wise MapTR. We use the 30-point Polyline to represent the path. We stack 6 deformable Transformer decoder layers and regress $(0,1)$ normalized coordinates. During training, $L_1$ loss and Focal loss are applied to the matched paths, and cross-entropy loss is applied to the auxiliary BEV segmentation branch. As for the Path2Graph, We discretize the path at the interval of $0.15m$, and the merging threshold is set to $0.15m$, the same as the graph interpolation size.




\subsection{Quantitative Comparision}
Tab.~\ref{tab:modeling} compares path-wise modeling with pixel-wise modeling and piece-wise modeling \wrt their accuracy, model size, FPS$_{\text{net}}$, FPS. 

\boldparagraph{Highlights.} Path-wise modeling achieves the best $F_1$ score on the both subgraph around the junction points and the overall graph across two popular datasets (nuScenes and Argoverse2), while running at the fastest inference speed. The designed Path2Graph algorithm has negligible cost in translating predicted paths into a directed lane graph (from 16.5 FPS to 15.6 FPS). 


\boldparagraph{Path-wise \vs pixel-wise.}
Thanks to the high-resolution output of the segmentation model, pixel-wise modeling exhibits comparable precision for subgraph construction and overall graph construction. Nevertheless, the notorious fragmentation and oversmooth issues of segmentation methods make it hard to maintain continuity across pixels and distinguish the fine-grained subgraph around the junction points, leading to much lower recall compared to path-wise modeling. 
In comparison, path-wise modeling demonstrates higher lane graph construction quality and 2x faster inference speed.

\boldparagraph{Path-wise \vs piece-wise.} Path-wise modeling outperforms piece-wise modeling on all metrics. Even equipped with the same advanced Transformer decoder proposed in \cite{maptr}, our path-wise LaneGAP still significantly outperforms piece-wise MapTR, which validates the effectiveness.


\subsection{Quantitative Comparision on Openlane-V2 Dataset}

Tab.~\ref{tab:toponet} compare path-wise modeling against state-of-the-art piece-wise method TopoNet on OpenLane-V2~\cite{openlanev2} dataset, which is a large-scale perception and reasoning dataset developed from nuScenes and Argoverse2. To further validate the superiority of path-wise modeling over piece-wise modeling under fair setting, we adapt the lane piece detection part of TopoNet for learning path-wise representation, while removing all the related piece-wise connection modules of TopoNet. This variant is denoted as LaneGAP$^{\dagger}_{\text{path}}$, distinguishing it from the original implementation, TopoNet${_\text{piece}}$. All other hyper-parameters and the training recipe remain consistent with TopoNet$_{\text{piece}}$.

\boldparagraph{Results.} Since the evaluation  metric proposed in TopoNet is not applicable to other modelings. We employ the mIoU metric for comparison, which is also utilized in TopoNet paper for benchmarking against the pixel-wise HDMapNet. As shown in Tab.~\ref{tab:toponet}, LaneGAP$^{\dagger}$ achieves 1.6 higher mIoU than TopoNet$_{\text{piece}}$ in $50m \times 100m$ perception range, thereby validating the effectiveness of path-wise modeling.

\subsection{Qualitative Comparison}

Fig.~\ref{fig:qualitative_comparison} compares baselines on complicated lane graphs with more than 6 junction points across different datasets. Failure case studies and more qualitative comparisons are provided in the supplementary material.

\begin{table*}[t!]
    \centering
    \begin{minipage}[t]{0.48\linewidth}
        \centering
        \renewcommand\arraystretch{1.1}
        \renewcommand\tabcolsep{3pt}
        \small
        \scalebox{0.62}{
        \begin{tabular}{l  c  c  c  c  c  c c c}
            \toprule
            \multirow{2}{*}{$\mathcal{L}_{\rm auxseg}$} & \multicolumn{3}{c}{Junction TOPO} & \multicolumn{3}{c}{TOPO} & \multirow{2}{*}{Param.} & \multirow{2}{*}{FPS}\\ 
             & Prec. & Rec. & \cellcolor{gray!20}$F_1$  & Prec. & Rec. &\cellcolor{gray!20}$F_1$ & &  \\
            \midrule
            \xmark &  0.544 & 0.516 &\cellcolor{gray!20}0.529 & 0.421 & 0.484 &\cellcolor{gray!20}0.450 & 35.9M & 15.6 \\
    
            \cmark &  0.554 & 0.524 &\cellcolor{gray!20}0.539 & 0.432 & 0.493 &\cellcolor{gray!20}0.461 & 35.9M & 15.6 \\
            \bottomrule
        \end{tabular}}
        \caption{Auxiliary BEV segmentation branch.}
        \label{tab:bevseg}
    \end{minipage}
    \hfill
    \begin{minipage}[t]{0.48\linewidth}
        \centering
        \renewcommand\arraystretch{0.9}
        \renewcommand\tabcolsep{3pt}
        \small
        \scalebox{0.62}{
            \begin{tabular}{l c  c  c  c  c  c c c}
                \toprule
                \multirow{2}{*}{\shortstack[l]{Polyline\\Points}} &  \multicolumn{3}{c}{Junction TOPO} & \multicolumn{3}{c}{TOPO} & \multirow{2}{*}{Param.} & \multirow{2}{*}{FPS}\\ 
                &  Prec. & Rec. & \cellcolor{gray!20}$F_1$  & Prec. & Rec. &\cellcolor{gray!20}$F_1$  \\
                \midrule
                10   & 0.550 & 0.515 & \cellcolor{gray!20}0.532 & 0.423 & 0.481 & \cellcolor{gray!20}0.450  & 35.9M & 16.3\\
                30   & 0.554 & 0.524 &\cellcolor{gray!20}0.539 & 0.432 & 0.493 &\cellcolor{gray!20}0.461  & 35.9M & 15.6\\
                40  & 0.566 & 0.548 &\cellcolor{gray!20}0.557 & 0.440 & 0.509 & \cellcolor{gray!20}0.472  & 35.9M & 15.1\\
                \bottomrule
            \end{tabular}}

        \caption{Polyline path representation.}
        \label{tab:polyline}
\end{minipage}
\end{table*}
\begin{table*}[t!]
    \centering
    \begin{minipage}[t]{0.48\linewidth}
        \centering
        \renewcommand\arraystretch{1.1}
        \renewcommand\tabcolsep{3pt}
        \small
        \scalebox{0.62}{
            \begin{tabular}{l  c  c  c  c  c  c c c}
                \toprule
                \multirow{2}{*} {\shortstack[l]{Bezier \\ Points}} &  \multicolumn{3}{c}{Junction TOPO} & \multicolumn{3}{c}{TOPO}& \multirow{2}{*}{Param.} & \multirow{2}{*}{FPS}\\ 
                & Prec. & Rec. &\cellcolor{gray!20}$F_1$  & Prec. & Rec. & \cellcolor{gray!20}$F_1$  \\
                \midrule
                3  & 0.480 & 0.483 & \cellcolor{gray!20}0.481 & 0.309 & 0.447 & \cellcolor{gray!20}0.365 
                &35.9M & 16.1\\
                5  & 0.514 & 0.507 &\cellcolor{gray!20}0.510 &0.375 & 0.477 & \cellcolor{gray!20}0.477  
                & 35.9M & 15.8\\
                10  & 0.498 & 0.503 & \cellcolor{gray!20}0.501 & 0.334 & 0.463 & \cellcolor{gray!20}0.388  & 35.9M & 15.6\\
                \bottomrule
            \end{tabular}}
        \caption{Bezier path representation.}
        \label{tab:bezier}
    \end{minipage}
    \hfill
    \begin{minipage}[t]{0.48\textwidth}
        \centering
        \renewcommand\arraystretch{1.1}
        \renewcommand\tabcolsep{2pt}
        \small
        \scalebox{0.58}{
            \begin{tabular}{l|cccc|cccc}
                \toprule
                \multirow{2}{*}{Method} &
                \multicolumn{4}{c|}{L2 (m) $\downarrow$} & 
                \multicolumn{4}{c}{Collision (\%) $\downarrow$} \\
                & 1s & 2s & 3s & \cellcolor{gray!30}Avg. & 1s & 2s & 3s & \cellcolor{gray!30}Avg.\\
                \midrule
                VAD-Tiny &0.46 &0.76& 1.12& \cellcolor{gray!30}0.78&0.21& 0.35& 0.58& \cellcolor{gray!30}0.38\\ 
                VAD-Tiny+MapTR& 0.43 & 0.74 & 1.06 & \cellcolor{gray!30}0.74 & 0.20 & 0.31 & 0.51 & \cellcolor{gray!30}0.34 \\
                VAD-Tiny+LaneGAP&\textbf{0.42}& \textbf{0.70} &\textbf{0.99}&\cellcolor{gray!30}\textbf{0.70} & \textbf{0.18} & \textbf{0.27} & \textbf{0.43} & \cellcolor{gray!30}\textbf{0.29} \\
                \bottomrule
            \end{tabular}}

    \caption{Comparison on downstream end-to-end planner VAD.}
    \label{tab:plan}
    \end{minipage}

\end{table*}

\begin{table*}[t!]
    \centering
    \begin{minipage}[t]{0.48\linewidth}
        \centering
        \renewcommand\arraystretch{1.1}
        \renewcommand\tabcolsep{3pt}
        \small
        \scalebox{0.62}{
        \begin{tabular}{l c  c  c  c  c  c  c }
            \toprule
            \multirow{2}{*}{Method} & \multirow{3}{*}{Modeling} & \multicolumn{3}{c}{Junction TOPO} & \multicolumn{3}{c}{TOPO} \\ 
            & & Prec. & Rec. & \cellcolor{gray!20}$F_1$  & Prec. & Rec. &\cellcolor{gray!20}$F_1$  \\
            \midrule
            HDMapNet &Pixel & 0.401&0.328&\cellcolor{gray!20}0.361&0.415&0.349&\cellcolor{gray!20}0.379 \\
            STSU &Piece & 0.309&0.297&\cellcolor{gray!20}0.303&0.298&0.273&\cellcolor{gray!20}0.285\\
            MapTR & Piece&0.369&0.357&\cellcolor{gray!20}0.363&0.363&0.317&\cellcolor{gray!20}0.338\\
            LaneGAP & Path&0.384&0.400&\cellcolor{gray!20}\textbf{0.392}&0.384&0.392&\cellcolor{gray!20}\textbf{0.388}\\
            \bottomrule
        \end{tabular}}

        \caption{Comparing on large perception range $50m \times 100m$.}
        \label{tab:perception_range}
    \end{minipage}
    \hfill
    \begin{minipage}[t]{0.48\linewidth}
        \centering
        \renewcommand\arraystretch{1.1}
        \renewcommand\tabcolsep{3pt}
        \small
        \scalebox{0.62}{
            \begin{tabular}{l c  c  c  c  c  c  c }
                \toprule
                \multirow{2}{*}{Method} & \multirow{3}{*}{Modeling} & \multicolumn{3}{c}{Junction TOPO} & \multicolumn{3}{c}{TOPO} \\ 
                & & Prec. & Rec. & \cellcolor{gray!20}$F_1$  & Prec. & Rec. &\cellcolor{gray!20}$F_1$  \\
                \midrule
                HDMapNet &Pixel &0.464&0.312&\cellcolor{gray!20}0.373&0.468&0.318&\cellcolor{gray!20}0.379\\
                STSU &Piece & 0.318&0.241&\cellcolor{gray!20}0.274&0.296&0.245&\cellcolor{gray!20}0.268\\
                MapTR & Piece&0.391&0.298&\cellcolor{gray!20}0.338&0.369&0.304&\cellcolor{gray!20}0.334\\
                LaneGAP & Path&0.458&0.356&\cellcolor{gray!20}\textbf{0.400}&0.447&0.358&\cellcolor{gray!20}\textbf{0.398}\\
                \bottomrule
            \end{tabular}}

        \caption{Comparing on new split.}
        \label{tab:newsplit}
    \end{minipage}

    \end{table*}
\begin{table*}[t!]
    \centering
    \begin{minipage}[t!]{0.48\linewidth}
        \centering
        \renewcommand\arraystretch{1.1}
        \renewcommand\tabcolsep{6pt}
        \small
        \scalebox{0.62}{
        \begin{tabular}{l|cccc}
            \toprule
                &HDMapNet & STSU & MapTR & LaneGAP \\
            \midrule
            mIoU on nusc &  48.1 & 38.2 & 43.6 & \textbf{49.2} \\
            mIoU on av2 &55.0&44.7&49.3&\textbf{55.9}\\
            \bottomrule
        \end{tabular}}
        \caption{Comparing on mIoU metric.}
        \label{tab:iou}
    \end{minipage}
    \hfill
    \begin{minipage}[t!]{0.48\linewidth}
        \centering
        \renewcommand\arraystretch{1.1}
        \renewcommand\tabcolsep{7.2pt}
        \small
        \scalebox{0.62}{
            \begin{tabular}{l  c  c  c  c  c  c c c}
                \toprule
                \multirow{2}{*} {\shortstack[l]{Post- \\ Proc.}} &  \multicolumn{3}{c}{Junction TOPO} & \multicolumn{3}{c}{TOPO} \\ 
                & Prec. & Rec. &\cellcolor{gray!20}$F_1$  & Prec. & Rec. & \cellcolor{gray!20}$F_1$ & \\
                \midrule
                Pixel  & 0.898 & 0.678 & \cellcolor{gray!20}0.773 & 0.913 & 0.693 & \cellcolor{gray!20}0.788 \\
                Piece  & 1.000 & 1.000 &\cellcolor{gray!20}1.000 & 1.000 & 1.000 & \cellcolor{gray!20}1.000 \\
                Path  & 1.000 & 1.000 & \cellcolor{gray!20}1.000 & 1.000 & 1.000 & \cellcolor{gray!20}1.000 \\
                \bottomrule
            \end{tabular}}

        \caption{Post-processing accuracy.}
        \label{tab:postproc}
    \end{minipage}

    \end{table*}

\boldparagraph{Highlights.} Path-wise modeling demonstrates better lane graph construction quality than pixel-wise and piece-wise modeling on extremely challenging lane graphs, well preserving the continuity of lane.

\boldparagraph{Path-wise \vs pixel-wise.} As shown in Fig.~\ref{fig:qualitative_comparison}, the pixel-wise HDMapNet performs well on non-junction area. But for the subgraph around the junction points, the segmentation model struggles to distinguish the fine-grained topology, and the post-processing is prone to fail to generate a decent vectorized lane graph,  aligning with the much lower Junction TOPO metrics in Tab.~\ref{tab:modeling}. On the contrary, LaneGAP can capture the fine-grained topology.

\boldparagraph{Path-wise \vs piece-wise.} The quality of the lane graph constructed by piece-wise modeling depends on the accuracy of both piece detection and connectivity prediction. As shown in Fig.~\ref{fig:qualitative_comparison}, the piece-wise MapTR either detects jagged lane pieces or produces wrong connections, leading to an incomplete and inaccurate lane graph after Piece2Graph post-processing. In contrast, our proposed path-wise modeling encodes the connectivity between pieces into the continuous path representation, which is easy to learn and robust.

\subsection{Ablation Study}
We ablate the design choices with a 24-epoch training schedule of path-wise modeling on nuScenes by default without specification. And we report Junction TOPO and TOPO only on the directed lane graph.

\boldparagraph{Effectiveness of auxiliary BEV segmentation branch.} Tab.~\ref{tab:bevseg} shows the auxiliary BEV segmentation branch can improve the performance by 1\% $F_1$ of Junction TOPO and 1.1\% $F_1$ of TOPO without adding cost at inference. 

\boldparagraph{Polyline path representation.}
As shown in Tab.~\ref{tab:polyline}, the accuracy increases with adding more points for Polyline modeling under a 24-epoch schedule, while the inference speed decreases. We choose 30-point Polyline as default setting.

\boldparagraph{Bezier path representation.}
Tab.~\ref{tab:bezier} shows that adding the number of control points from 3 to 5 improves 2.9\% $F_1$ of Junction TOPO and 11.2\% $F_1$ of TOPO. While adding the number of control points from 5 to 10 incurs an accuracy drop. 

\boldparagraph{Comparison on end-to-end planner VAD.}
Tab.~\ref{tab:plan} provides comparison on end-to-end planner VAD~\cite{vad} by including an extra centerline lane graph branch. Since VAD utilizes instance-wise queries, HDMapNet is not applicable, we only compare with piece-wise MapTR. Path-wise modeling enhances planning performance and outperforms piece-wise modeling. Notably, LaneGAP’s path queries offer a complete depiction of the lane graph, unlike MapTR's centerline piece queries, whose connectivity predictions are not directly usable in VAD.








\boldparagraph{Large perception range.} In Tab.~\ref{tab:perception_range}, we compare the performance on a large perception range ($50m \times 100m$) using the same training recipe of Tab.~\ref{tab:modeling}. LaneGAP still outperforms other methods.

\boldparagraph{Generalization on new split.} StreamMapNet~\cite{streammapnet} proposes a new split of the nuScenes dataset, which splits the train/val set by the geolocation and reduces the map overlap. We further compare the performance on this new split in Tab.~\ref{tab:newsplit} using the same training recipe. LaneGAP maintains consistent superiority.

\boldparagraph{Comparison on mIoU metric.} In Tab.~\ref{tab:iou}, we further compare the performance of different modelings with mIoU metric following HDMapNet~\cite{hdmapnet}. We render the prediction on the BEV map with $0.75m$ line width. The results further validate the superiority of LaneGAP.

\boldparagraph{Post-processing accuracy.} In Tab.~\ref{tab:postproc}, we show the accuracy of the lane graphs transformed by respective post-processings given the ground-truth segmentation maps, sets of pieces and connections, and set of paths. The results indicate that Pixel2Graph post-processing is prone to fail in handling the fine-grained topology of the subgraph around junction points, as illustrated in 
 Fig.~\ref{fig:qualitative_comparison}.  
\label{para:postproc-acc}

        
 




\section{Conclusion}
In this work, we present an online lane graph construction method LaneGAP based on novel path-wise modeling. We qualitatively and quantitatively demonstrate the superiority of LaneGAP over pixel-based and piece-based methods. 
LaneGAP  can serve as a fundamental module of the self-driving 
system and facilitate downstream motion prediction and planning, which we leave as future work.

\section*{Acknowledgement}
This work was partly supported by the National Natural Science Foundation of China (NSFC 62276108).

%
%
\bibliographystyle{splncs04}
\bibliography{main}

\clearpage
\appendix

\section{Training Settings for Quantitative Comparison}
In Tab.~\ref{tab:nusc-config} and Tab.~\ref{tab:av2-config}, we summarize the detailed training settings for the quantitative comparison in the main paper. We use the same settings under the long schedule.

\vspace{-10pt}
\begin{table*}[h!]
    \centering
    \begin{minipage}[t]{0.48\linewidth}
        \centering
        \renewcommand\arraystretch{1.1}
        \renewcommand\tabcolsep{3pt}
        \small
        \scalebox{0.62}{
        \begin{tabular}{l  c  c  c  }
            \toprule
            Config & Pixel-wise & Piece-wise & Path-wise \\
            \midrule
            optimizer & \multicolumn{3}{c}{AdamW~\cite{loshchilov2017adamw}}\\
            optimizer hyper-parameter & \multicolumn{3}{c}{$\beta_1,\beta_2,\epsilon =  0.9, 0.999,$ 1e-8} \\
            gradient clip norm type & \multicolumn{3}{c}{2} \\
            gradient clip max norm & \multicolumn{3}{c}{35} \\
            learning rate & \multicolumn{3}{c}{6e-4}\\ 
            learning rate schedule  & \multicolumn{3}{c}{cosine decay~\cite{loshchilov2016cosineanneal}} \\
            warmup iteration & \multicolumn{3}{c}{500} \\
            weight decay & \multicolumn{3}{c}{0.01} \\
            \midrule
            input resolution & \multicolumn{3}{c}{$6\times900\times1600$} \\
            image resize ratio & \multicolumn{3}{c}{0.5} \\
            batch size & \multicolumn{3}{c}{32} \\
            training epochs & \multicolumn{3}{c}{110} \\
            BEV grid size & \multicolumn{3}{c}{$0.3m$} \\
                \bottomrule
            \end{tabular}}
        \caption{\textbf{Detailed training settings on nuScenes dataset.} We use the same training settings for quantitative comparison. }
        \label{tab:nusc-config}
    \end{minipage}
    \hfill
    \begin{minipage}[t]{0.48\linewidth}
        \centering
        \renewcommand\arraystretch{1.1}
        \renewcommand\tabcolsep{3pt}
        \small
        \scalebox{0.62}{
            \begin{tabular}{l  c  c  c  }
                \toprule
                Config & Pixel-wise & Piece-wise & Path-wise \\
                \midrule
                optimizer & \multicolumn{3}{c}{AdamW~\cite{loshchilov2017adamw}}\\
                optimizer hyper-parameter & \multicolumn{3}{c}{$\beta_1,\beta_2,\epsilon =  0.9, 0.999,$ 1e-8} \\
                gradient clip norm type & \multicolumn{3}{c}{2} \\
                gradient clip max norm & \multicolumn{3}{c}{35} \\
                learning rate & \multicolumn{3}{c}{6e-4}\\ 
                learning rate schedule  & \multicolumn{3}{c}{cosine decay~\cite{loshchilov2016cosineanneal}} \\
                warmup iteration & \multicolumn{3}{c}{500} \\
                weight decay & \multicolumn{3}{c}{0.01} \\
                \midrule
                input resolution & \multicolumn{3}{c}{$7\times2048\times2048$} \\
                image resize ratio & \multicolumn{3}{c}{0.3} \\
                batch size & \multicolumn{3}{c}{32} \\
                training epochs & \multicolumn{3}{c}{24} \\
                BEV grid size & \multicolumn{3}{c}{$0.3m$} \\
                \bottomrule
            \end{tabular}}
            \caption{\textbf{Detailed training settings on Argoverse2 dataset.} We use the same training settings for quantitative comparison. }
            \label{tab:av2-config}
\end{minipage}
\vspace{-30pt}
\end{table*}

\section{More Ablation Studies}

\boldparagraph{Modality.}
Tab.~\ref{tab:modality} shows that LiDAR modality builds a more accurate lane graph than vision modality (6.1\% higher $F_1$ of Junction TOPO and 4.8\% higher $F_1$ of TOPO). Fusing them further boosts performance.

\begin{table}[ht!]
    \centering
        \begin{tabular}{l  c  c  c  c  c  c c c }
            \toprule
            \multirow{2}{*}{Modality} & \multicolumn{3}{c}{Junction TOPO} & \multicolumn{3}{c}{TOPO} & \multirow{2}{*}{Param.} & \multirow{2}{*}{FPS}\\ 
             & Prec. & Rec. & \cellcolor{gray!20}$F_1$  & Prec. & Rec. &\cellcolor{gray!20}$F_1$  \\
            \midrule
            Vision-only &  0.554 & 0.524 &\cellcolor{gray!20}0.539 & 0.432 & 0.493 &\cellcolor{gray!20}0.461
            & 35.9M & 15.6\\
    
            LiDAR-only &  0.608 & 0.594 &\cellcolor{gray!20}0.600 & 0.467 & 0.559 &\cellcolor{gray!20}0.509& 9.1M & 8.6\\
            V \& L  &0.630 & 0.600 &\cellcolor{gray!20}0.615 & 0.504 & 0.561 &\cellcolor{gray!20}0.531 & 39.8M & 5.5\\
            \bottomrule
        \end{tabular}
    \caption{\textbf{Ablation for modality.}}
    \label{tab:modality}
\vspace{-15pt}
\end{table}

\boldparagraph{Training schedule.}
Tab.~\ref{tab:schedule} shows that adding more epochs mainly increases the TOPO metric and benefits little Junction TOPO, especially the recall of Junction TOPO. The 110-epoch trained multi-modality experiment further pushes the performance of lane graph construction.

\begin{table}[h!]
    \centering

    \begin{tabular}{l c  c  c  c  c  c  c }
    \toprule
    \multirow{2}{*}{Modality} & \multirow{3}{*}{Epoch} & \multicolumn{3}{c}{Junction TOPO} & \multicolumn{3}{c}{TOPO} \\ 
    & & Prec. & Rec. & \cellcolor{gray!20}$F_1$  & Prec. & Rec. &\cellcolor{gray!20}$F_1$  \\
    \midrule
    Vision-only &24 &  0.554 & 0.524 &\cellcolor{gray!20}0.539 & 0.432 & 0.493 &\cellcolor{gray!20}0.461 \\
    Vision-only &110 & 0.591 & 0.539 &\cellcolor{gray!20}0.564 & 0.547 & 0.511 &\cellcolor{gray!20}0.529 \\
    \midrule
    Vision \& LiDAR  & 24 &0.630 & 0.600 &\cellcolor{gray!20}0.615 & 0.504 & 0.561 &\cellcolor{gray!20}0.531 \\
    Vision \& LiDAR  & 110&0.668 & 0.601 & \cellcolor{gray!20}0.632 & 0.620 & 0.576 & \cellcolor{gray!20}0.597 \\
    \bottomrule
\end{tabular}
\caption{\textbf{Ablation for training schedule.} }
\label{tab:schedule}
\vspace{-15pt}
\end{table}

\boldparagraph{Tradeoff Ccomparison of modelings.}
In Tab.~\ref{tab:cost}, we report the tradeoff comparison of modelings in terms of accuracy and cost on nuScenes dataset.
\begin{table}[ht!]
    \centering
    \begin{tabular}{l  c  c  c  c  r  c }
        \toprule
        Method & Modeling & Modality &$F_1$@Junction & $F_1$@TOPO & FPS & Training time \\
        \midrule
        HDMapNet & Pixel-wise & V & 0.507 & 0.513 & 6.8 & 42h \\
        MapTR & Piece-wise & V & 0.496 & 0.465 & 15.0 & 49h\\
        \midrule
        LaneGAP & Path-wise & V & 0.564 & 0.529 & 15.6 & 58h\\
        LaneGAP & Path-wise & V\&L &0.632 & 0.597 & 5.5 & 60h\\
        \bottomrule
    \end{tabular}
\vspace{2pt}
\caption{\textbf{Tradeoff comparison on nuScenes dataset.} ``V'' and ``L'' respectively denote Vision modality and LiDAR modality. ``$F_1$@Junction'' and ``$F_1$@TOPO'' respectively denote $F_1$ of Junction TOPO and $F_1$ of TOPO on directed graph.}
\label{tab:cost}
\vspace{-15pt}
\end{table}

\section{Visualizations for Ablations}
\begin{figure}[t!]
    \centering
    \includegraphics[width=0.9\linewidth]{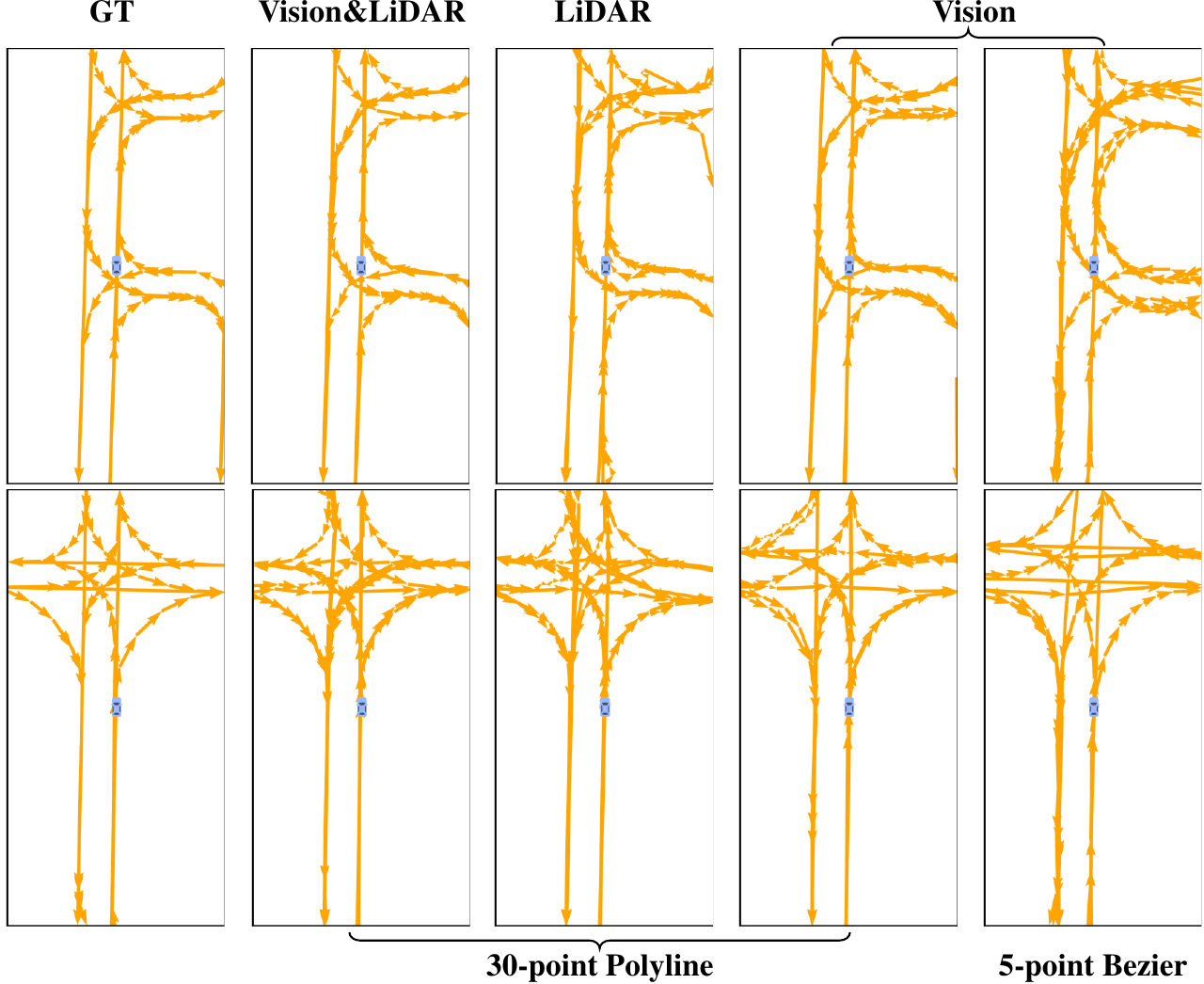}
    \caption{\textbf{Visualizations for ablations.} The visualized  ablation experiments are under a 24-epoch training schedule. The multi-modality method provides more accurate lane graph, and the Bezier method outputs smoother paths.}
    \label{fig:ablation_comp}
    \vspace{-4pt}
\end{figure}

In Fig.~\ref{fig:ablation_comp}, we visualize the 24-epoch trained ablation experiments: multi-modality in column 2, LiDAR modality in column 3, vision modality in column 4 and 5, 30-point polyline representation in column 2, 3 and 4, 5-point Bezier representation in column 5. As shown in Fig.~\ref{fig:ablation_comp}, the multi-modality method outputs a more accurate lane graph, and the Bezier method outputs smoother paths.

\begin{figure*}[t!]
    \centering
    \includegraphics[width=0.95\linewidth]{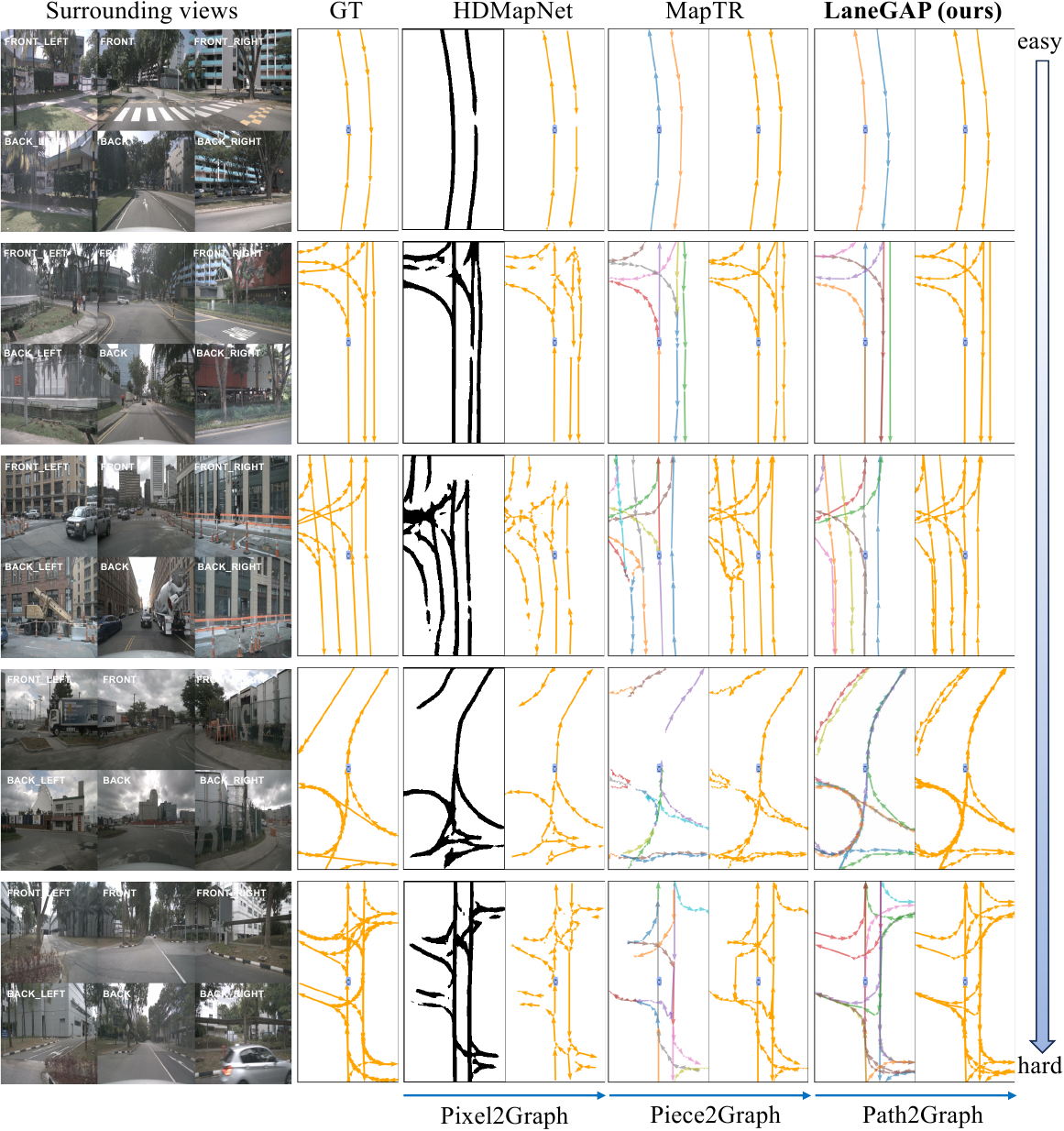}
    \caption{\textbf{Qualitative comparisons of increasing junction points.} We qualitatively compare path-wise LaneGAP with pixel-wise HDMapNet and piece-wise MapTR on lane graphs with increasing junction points. All models perform well on simple lane graphs with few junction points. However, as the number of junction points increases, HDMapNet and MapTR struggle to predict reasonable lane graphs, while our proposed LaneGAP continues to deliver robust results. }
    \label{fig:junctions}
    \vspace{-4pt}
\end{figure*}

\begin{figure}[th!]
    \centering
    \includegraphics[width=0.95\linewidth]{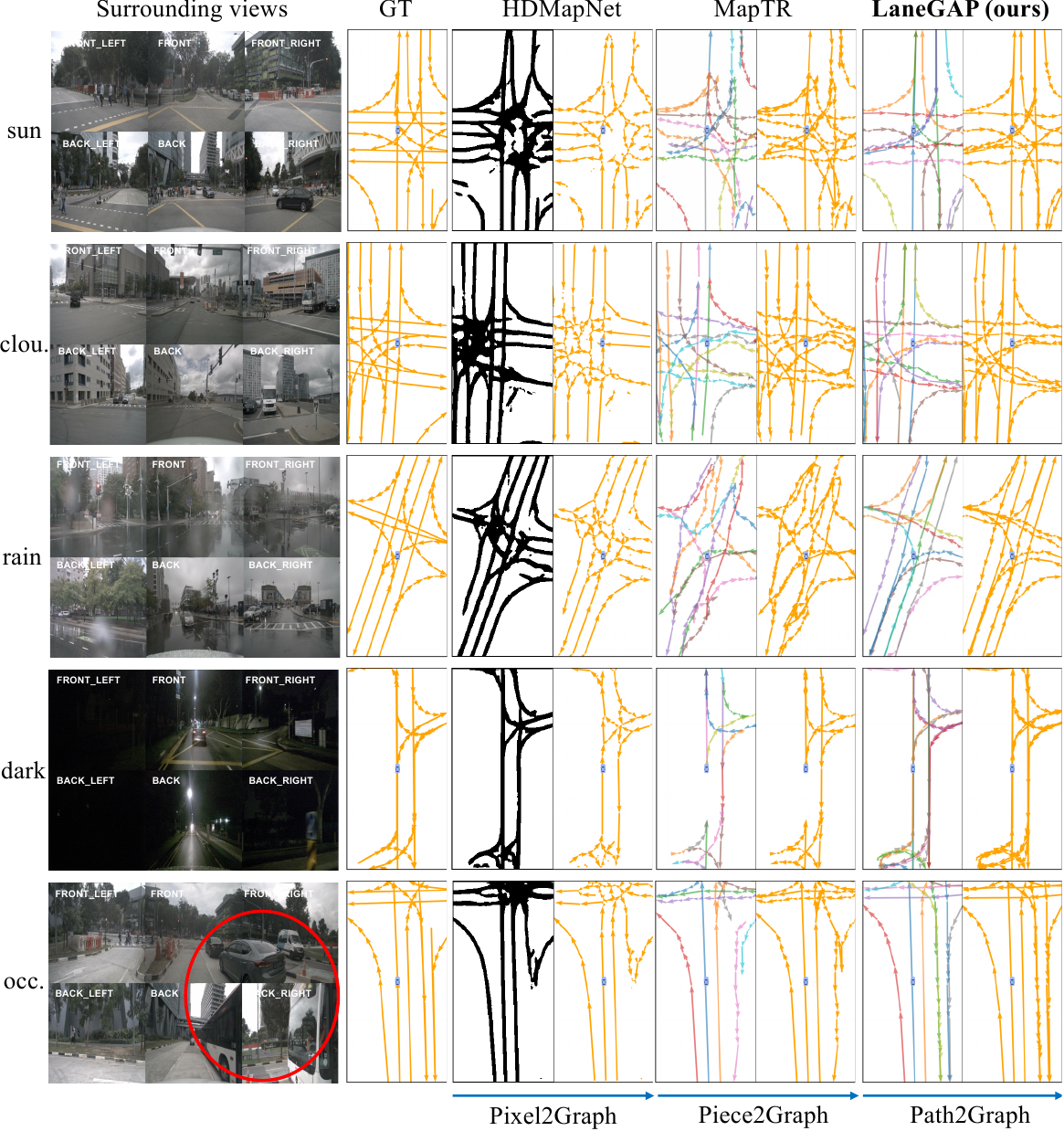}
    \caption{\textbf{Qualitative comparisons of diverse driving conditions.} We qualitatively compare path-wise LaneGAP with pixel-wise HDMapNet and piece-wise MapTR across sunny, cloudy, rainy, dark, and occluded driving conditions. LaneGAP consistently demonstrates superior performance across all conditions. Specifically, in the occluded condition (row 5), results indicate that pixel-wise and piece-wise modelings, which may break continuity, are prone to predict incomplete lane graphs. In contrast, our path-wise modeling approach preserves the continuity of the lane graph, enhancing robustness in occluded driving conditions.}
    \label{fig:conditions}
    \vspace{-4pt}
\end{figure}

\begin{figure}[th!]
    \centering
    \includegraphics[width=0.95\linewidth]{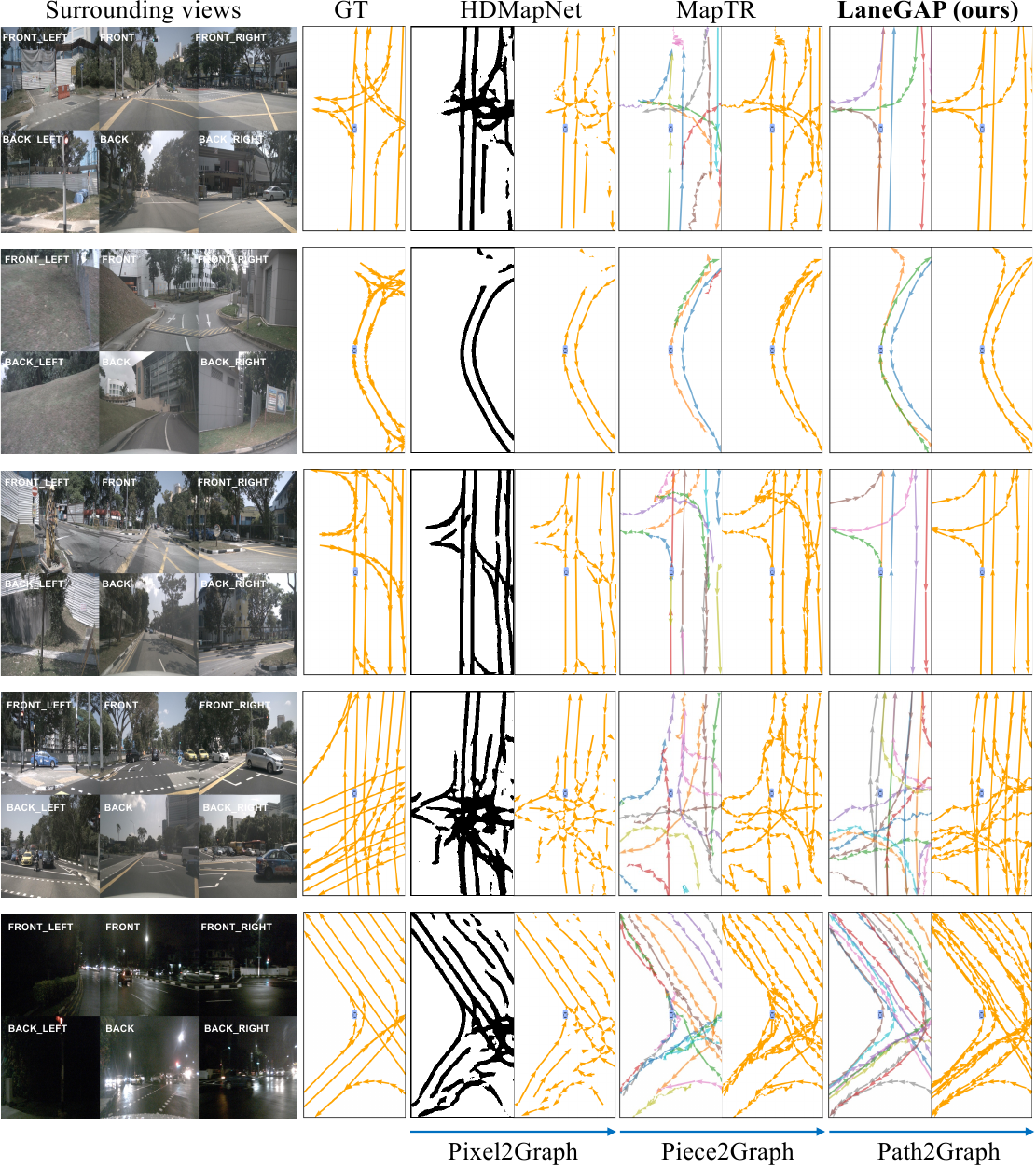}
    \caption{\textbf{Qualitative comparisons of failure cases.} We qualitatively compare path-wise LaneGAP with pixel-wise HDMapNet and piece-wise MapTR on the cases, where LaneGAP fails to handle. LaneGAP struggles to predict a complete subgraph for distant junction areas (row 1 and 2), located at the edge of the perception range. Furthermore, the extremely hard online driving conditions and topology (occlusion in row 4 and darkness in row 5) also make the construction very challenging.  Addressing these issues requires further research and development.}
    \label{fig:fail}
    \vspace{-4pt}
\end{figure}

\section{More Qualitative Comparisons}
\boldparagraph{Qualitative comparisons of increasing junction points.} In Fig.~\ref{fig:junctions}, we compare path-wise LaneGAP with pixel-wise HDMapNet, and piece-wise MapTR on lane graphs with increasing junction points. All models perform well on simple lane graphs with few junction points. However, as the number of junction points increases, HDMapNet and MapTR struggle to predict reasonable lane graphs, while our proposed LaneGAP continues to deliver robust results.

\boldparagraph{Qualitative comparisons of diverse driving conditions.} In Fig.~\ref{fig:conditions}, we compare these modelings across diverse driving conditions, including sunny, cloudy, rainy, dark, and occluded scenarios.  LaneGAP consistently demonstrates superior performance across all conditions. Specifically, in the occluded condition (row 5), results indicate that pixel-wise and piece-wise modelings, which may break continuity, are prone to predict incomplete lane graphs. In contrast, our path-wise modeling approach preserves the continuity of the lane graph, enhancing robustness in occluded driving conditions.

\boldparagraph{Qualitative comparisons of failure cases.}
In Fig.~\ref{fig:fail}, we compare these modelings on the cases, where LaneGAP fails to handle. The results show that LaneGAP struggles to predict a complete subgraph for distant junction areas (row 1 and 2), located at the edge of the perception range. Furthermore, the extremely hard online driving conditions and topology (occlusion in row 4 and darkness in row 5) also make the construction very challenging.  Addressing these issues requires further research and development.

\section{Video Visualizations}
We visualize the whole nuScenes validation set and attach videos to the supplementary material. The videos are generated with two models: the path-wise method based on only vision input and the path-wise method based on vision and LiDAR input.


\end{document}